\documentclass[10pt,twocolumn,letterpaper]{article}

  \usepackage[pagenumbers]{cvpr} %

\usepackage{multirow}
\usepackage{colortbl}
\usepackage{tabularx}
\newcolumntype{Y}{>{\centering\arraybackslash}X}
\definecolor{ourrow}{HTML}{EDEAF5}
\definecolor{oursbar}{HTML}{623F99}
\newcommand{\oursbar}{\rlap{\hspace*{-\tabcolsep}\textcolor{oursbar}{\rule[\dimexpr-\dp\strutbox+0.4pt\relax]{1.8pt}{\dimexpr\ht\strutbox+\dp\strutbox-0.8pt\relax}}}}
\newcommand{\best}[1]{\textbf{#1}}
\newcommand{\second}[1]{\underline{#1}}
\newcommand{\oracle}[1]{\textcolor{gray}{#1}}

\makeatletter\def\fvm@Scale{0.92}\makeatother
\newcommand{\mname}[1]{{\fontencoding{T1}\fontfamily{fvm}\fontseries{m}\selectfont #1}}

\usepackage{tikz}
\definecolor{iconamber}{HTML}{E8900F}
\definecolor{iconviolet}{HTML}{8A63D2}
\definecolor{iconblue}{HTML}{3B82F6}
\definecolor{iconvermillion}{HTML}{E4572E}
\definecolor{iconteal}{HTML}{2A9D8F}
\DeclareRobustCommand{\camicon}[1]{\hspace{-0.11em}\ensuremath{\vcenter{\hbox{\tikz[x=1em, y=1em, line width=0.45pt,
  line join=round, line cap=round]{%
  \filldraw[draw=#1, fill=#1!20] (0.5,-0.02) -- (1.02,0.1) -- (1.02,0.86) -- (0.5,0.74) -- cycle;
  \draw[#1] (0,0.42) -- (0.5,-0.02) (0,0.42) -- (0.5,0.74) (0,0.42) -- (1.02,0.1) (0,0.42) -- (1.02,0.86);}}}}\hspace{0.11em}}
\DeclareRobustCommand{\pathicon}{\hspace{-0.08em}\ensuremath{\vcenter{\hbox{\tikz[x=1em, y=1em, line width=0.8pt,
  line cap=round]{%
  \draw[iconblue] (0,0.36) .. controls (0.5,0.38) and (0.75,0.70) .. (1.15,0.72);
  \draw[iconteal] (0,0.36) .. controls (0.5,0.34) and (0.75,0.02) .. (1.15,0.0);
  \draw[iconvermillion] (0,0.36) -- (1.2,0.36);}}}}\hspace{0.08em}}

\definecolor{cvprblue}{rgb}{0.21,0.49,0.74}
\usepackage[pagebackref,breaklinks,colorlinks,allcolors=cvprblue]{hyperref}

\usepackage{xcolor}
  \makeatletter
  \def\Float@macros#1{#1\@addtodblcol#1\@addtocurcol#1\@addtotoporbot#1\@addtobot#1\@doclearpage#1\@combinedblfloats}
  \def\Float@savekernel#1{\expandafter\let\csname kernel\string#1\endcsname#1}
  \def\Float@savest#1{\expandafter\let\csname stfloats\string#1\endcsname#1}
  \def\Float@usekernel#1{\expandafter\let\expandafter#1\csname kernel\string#1\endcsname}
  \def\Float@usest#1{\expandafter\global\expandafter\let\expandafter#1\csname stfloats\string#1\endcsname}
  \Float@macros\Float@savekernel
  \usepackage{stfloats}
  \Float@macros\Float@savest
  \Float@macros\Float@usekernel
  \newcommand{\SuppFloats}{\Float@macros\Float@usest}
  \makeatother

\def\confName{CVPR}
\def\confYear{2027}

\title{TrackCast: Forecasting Object Motion as Full-Surface 3D Point Tracks}

\author{Jacob Lin \qquad Simon Weber \qquad Ronald Clark\\
Department of Computer Science, University of Oxford
}

\begin{document}
\twocolumn[{%
\renewcommand\twocolumn[1][]{#1}%
\maketitle
\vspace{-16pt}\begin{center}
  \includegraphics[width=0.95\linewidth]{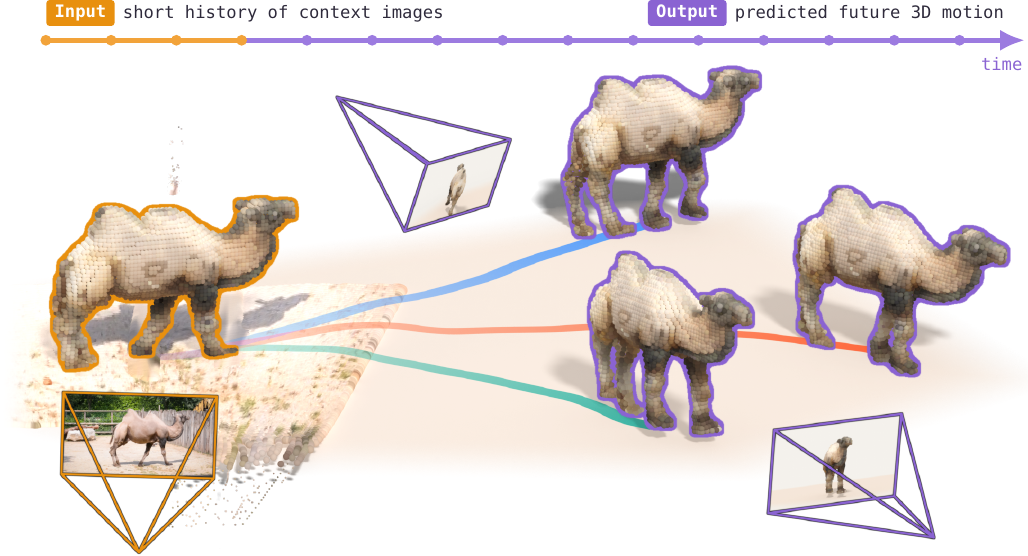}
  \captionof{figure}{\textbf{TrackCast} forecasts the full-surface 3D motion of an object from a short history of context images~(\camicon{iconamber}), producing predictions that can be viewed from any angle~(\camicon{iconviolet}) and steered toward different futures~(\pathicon). Trained on in-the-wild data from our automatic pipeline, it parameterizes motion as dense 3D point tracks, enabling token-efficient generation at one token per track.}
  \label{fig:banner}
\end{center}
}]
\begin{abstract}
Anticipating how objects move is a fundamental capability of visual intelligence.
Point tracks describe this motion independently of appearance, but most point-track forecasters either predict 2D trajectories of in-view pixels or lift those pixels to 3D with per-pixel depths. They track pixels, not objects, and therefore forecast only the visible part of an object's surface.
We introduce the task of full-surface 3D point-track forecasting: predicting the future 3D tracks of every point on an object's surface, seen and unseen, from a short history of frames.
These tracks cover the whole object, so they can be expressed in an object-centric canonical space.
To leverage this, we present TrackCast, a flow-matching diffusion transformer that takes the tracks reconstructed from the context frames and forecasts every track jointly in the canonical space, which captures the object's deformation, and in world space, which captures its global motion.
It can be conditioned on how much the object moves in world space and deforms in canonical space.
Since no real-world dataset provides such tracks for training, we build an automatic pipeline that extracts them from in-the-wild videos, and release AiM-Track, over 7k full-surface track sequences of animals.
On three animal datasets and on robotic manipulation of deformable objects, TrackCast forecasts deformation more accurately than state-of-the-art 3D forecasters trained on the same full-surface tracks.
\\Project page: {\urlstyle{same}\url{https://trackcast.jacob-lin.com/}}\end{abstract}
    
\section{Introduction}
\label{sec:intro}

Forecasting object motion has attracted growing attention in 2D, both by generating future video frames~\cite{video_diffusion_models,blattmann2023stablevideodiffusionscaling} and by predicting future trajectories of image points~\cite{WenLS0D0A24,track2act,fmitw}.
Point tracks provide an appearance-independent representation of motion~\cite{fmitw} and have been adopted for robotic manipulation~\cite{WenLS0D0A24,track2act}, physical scenes~\cite{boduljak2026what}, general videos~\cite{Lu_2026_CVPR}, and animals in the wild~\cite{fmitw}.
However, image-plane trajectories cannot fully capture 3D motion: movement toward the camera may produce little image displacement, while out-of-plane rotations can appear as object deformation.

Recent 3D forecasters such as MolmoMotion~\cite{zhang2026molmomotion} and PointWorld~\cite{point_world} address this limitation by lifting pixels to 3D with per-pixel depth.
While this enables forecasting motion in 3D, it does not overcome the fundamental challenge of partial visibility: cameras observe only part of an object's surface at a time.
As a result, these methods forecast only the motion of the \textit{visible} surface.
This limits their use in robotics, animation, and augmented reality, where the motion of the entire object matters.
Current point-track forecasters trained on real data thus track pixels, not objects.

We introduce the task of \textit{full-surface 3D point-track forecasting}: from a short history of frames, we predict the future 3D tracks of every point on an object's surface, \textit{both} seen and unseen.
These tracks cover the whole object, so they can be expressed in an object-centric canonical space.
To leverage this, we present TrackCast, a flow-matching~\cite{flow_matching} diffusion transformer~\cite{diffusion_transformer} that forecasts every track jointly in this space, which captures the object's deformation, and in world space, which captures its global motion.
It can be conditioned on how much the object moves in world space and deforms in canonical space.
The 3D forecasts can also be rendered from novel viewpoints (\cref{fig:banner}).

Training such a model, however, requires full-surface 3D tracks of real objects in motion, which no existing dataset provides.
To address this gap, we build an automatic pipeline that reconstructs the complete object geometry in every frame and extracts temporally consistent 3D tracks of its whole surface, seen and unseen.
Applying this pipeline to the animal videos of AiM~\cite{zhao2025animalvideo}, we construct and release AiM-Track, a training set and benchmark of full-surface 3D tracks with over 7k training sequences of 23 animal categories and a held-out test split.

In summary, our contributions are:
\begin{itemize}
  \item We introduce the task of full-surface 3D point-track forecasting, which extends forecasting from the visible surface to the whole object, and release AiM-Track, over 7k full-surface track sequences that our automatic pipeline extracts from in-the-wild animal videos.
  \item We propose TrackCast, a flow-matching diffusion transformer designed for full-surface tracks, which jointly forecasts an object's deformation in its canonical space and its global motion in world space, with optional conditioning on how much the object moves and deforms.
  \item We show that, on three animal datasets and on robotic manipulation of deformable objects, TrackCast forecasts deformation more accurately than state-of-the-art 3D forecasters trained on the same full-surface tracks. An ablation shows that the canonical space lowers the canonical error on every animal dataset and the world error on DT4D~\cite{dt4d} and DAVIS~\cite{pont2017davis}.
\end{itemize}

\section{Related Work}
\label{sec:related_works}

\paragraph{Track forecasting.}
Forecasting point tracks instead of pixels~\cite{blattmann2023stablevideodiffusionscaling,video_diffusion_models} separates an object's motion from its appearance~\cite{fmitw}.
Most existing methods forecast tracks in 2D~\cite{xu2025flow,Chen_2025_ICCV,gao2025flip,Yang_2025_CVPR,li2024_GenerativeImageDynamics}.
ATM~\cite{WenLS0D0A24} and Track2Act~\cite{track2act} do so for robotic manipulation from an image of the scene, conditioned on a language instruction and a goal image, respectively.
Boduljak et al.~\cite{boduljak2026what} generate dense 2D trajectories from a single image, and Lu et al.~\cite{Lu_2026_CVPR} track and forecast points with a video-conditioned flow-matching model.
Closest to our setting, Thakkar et al.~\cite{fmitw} forecast tracks of animals in the wild with a diffusion transformer, but in 2D.

A few recent methods forecast tracks in 3D~\cite{zhang2026molmomotion,point_world,pmlr-v270-yuan25a,niu2025pretraining,bharadhwaj2026motionforesight,3pointr}.
MolmoMotion~\cite{zhang2026molmomotion} forecasts the trajectories of query points on an object, conditioned on a language description of the action, and PointWorld~\cite{point_world} predicts the 3D point flow of a scene, conditioned on robot actions.
On real videos, both obtain their 3D tracks by lifting visible 2D tracks~\cite{Harley_2025_ICCV,Karaev_2025_ICCV} with estimated depth, so, like the 2D methods, they are restricted to points the camera sees.
3DPWM~\cite{peri2026pwm} instead completes partial point clouds before forecasting their motion under robot actions, and is trained only on simulated tabletop scenes.

In contrast, we learn to forecast 3D tracks of an object's full surface from in-the-wild videos.

\paragraph{Generative 3D and 4D reconstruction.}
Image-to-3D models predict a complete, plausible 3D shape of an object, including its unseen parts, from a single image~\cite{Xiang_2025_CVPR,Xiang_2026_CVPR,hunyuan3d22025tencent,sam3d}.
SAM 3D~\cite{sam3d} does so for objects in natural images and also predicts the object's pose.

Generative 4D reconstruction methods recover moving objects from a monocular video.
L4GM~\cite{ren2024l4gm} predicts per-frame 3D Gaussians, and ActionMesh~\cite{Sabathier_2026_CVPR} predicts an animated mesh of fixed topology, both without per-video optimization.
Lift4D~\cite{litman2026lift4d} makes per-frame SAM 3D reconstructions temporally consistent, fits a single deformable 3D Gaussian representation to them and to the video, and completes unobserved appearance with a diffusion prior.

These generative reconstructions cover an object's full surface, including the parts the camera does not see.
Building on SAM 3D and Lift4D, we reconstruct objects from in-the-wild videos frame by frame and track every surface point, yielding full-surface 3D point tracks for training.

\paragraph{Articulated motion modeling.}
Keypoints~\cite{Mathisetal2018} and parametric body models such as SMPL~\cite{SMPL:2015} and SMAL~\cite{Zuffi:CVPR:2017} are common representations of articulated motion, and MagicPony~\cite{wu2023magicpony} and 3D-Fauna~\cite{li2024fauna} learn articulated animal models from in-the-wild images.
Ponymation~\cite{sun2024ponymation} and Kirin~\cite{zhao2026kirin}, which reconstructs SMAL motions from the AiM videos~\cite{zhao2025animalvideo}, generate the motion of such models rather than forecast an observed one, and human motion prediction forecasts future skeletal poses or body meshes~\cite{Martinez_2017_CVPR,chen2023humanmac,Zhang_2019_ICCV}.
All of these rely on keypoints, a skeleton, or a template defined in advance for a class of objects.

In contrast, we forecast surface tracks, which need no such predefined structure, so the same approach applies to animals and to objects in robotic manipulation.

\section{Full-Surface 3D Point Tracks}
\label{sec:data_processing}

\begin{figure*}[t]
  \centering
  \includegraphics[width=0.92\linewidth]{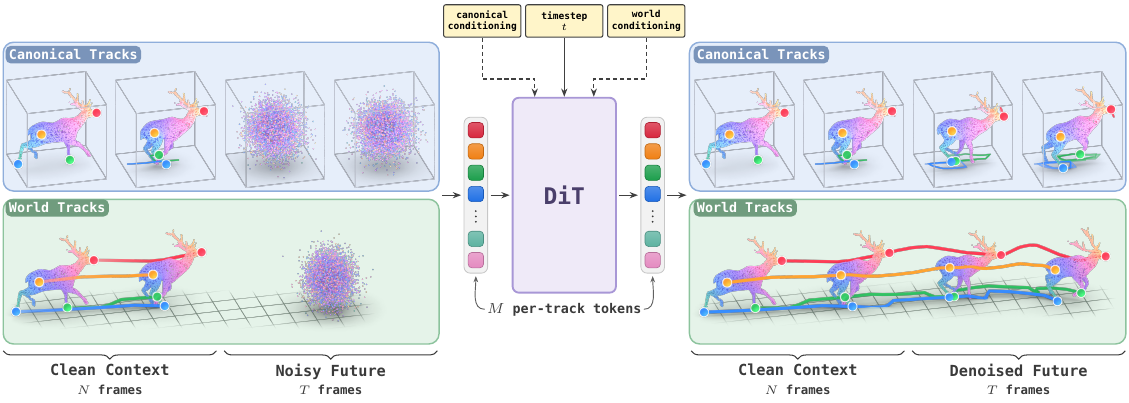}
  \caption{\textbf{Method overview.} TrackCast forecasts an object's motion as $M$ full-surface 3D point tracks, each expressed in two spaces: an object-centric canonical space that isolates deformation, and a world space that adds rigid motion. One token per track holds its clean context and noisy future in both spaces, and a diffusion transformer (DiT) jointly denoises both futures by flow matching. The DiT can optionally be conditioned on world and canonical motion magnitudes, which control how much the object moves and how much it deforms (\cref{fig:control_behaviour} and \cref{fig:control_gallery}, top).}
  \label{fig:method_overview}
  \vspace{-8pt}
\end{figure*}

We represent an object's motion as 3D point tracks sampled over its entire surface, including regions that are not visible in the input images.
Our goal is to obtain temporally consistent tracks in two coordinate systems: a canonical space describing the object's deformation, and a world space additionally accounting for its rigid motion.
We introduce an automatic pipeline to extract such tracks from in-the-wild videos (\cref{sec:data_real}), which we apply to the animal videos of AiM~\cite{zhao2025animalvideo} to construct AiM-Track.
We also obtain tracks directly from animated meshes for additional synthetic training data (\cref{sec:data_synthetic}).
These tracks serve as both inputs and training targets for our forecasting model (\cref{sec:method}).

\subsection{Extraction from In-the-Wild Videos}
\label{sec:data_real}

\paragraph{Temporally consistent 3D reconstruction.}
Given a video of $F$ frames and corresponding object masks, we reconstruct the object in each frame using SAM 3D~\cite{sam3d}.
For each frame $n$, SAM 3D predicts the object's complete shape in canonical coordinates, together with its pose, and we take the outer surface of this shape as $\mathcal{S}_n$.
However, SAM 3D reconstructs each frame independently from a single image, so its reconstructions can be temporally inconsistent in shape and orientation.

To improve consistency, we follow Lift4D~\cite{litman2026lift4d} and initialize each frame's generation from the preceding reconstruction.
Specifically, SAM 3D generates the reconstruction of frame $n$ by integrating a flow-matching process over a latent variable $\mathbf{Z}^{n}_{t}$, where $t \in [0, 1]$ denotes the flow time, with $t = 0$ corresponding to noise and $t = 1$ to a clean latent.
For the first frame, we sample $\mathbf{Z}^{0}_{0} \sim \mathcal{N}(\mathbf{0}, \mathbf{I})$.
For subsequent frames $n > 0$, we initialize the process at $t_0 \in (0, 1)$ using
\begin{equation}
  \mathbf{Z}^{n}_{t_0} = (1 - t_0)\,\mathbf{Z}^{0}_{0} + t_0\,\mathbf{Z}^{n-1}_{1},
  \label{eq:causal_init}
\end{equation}
where $\mathbf{Z}^{n-1}_{1}$ is the clean latent reconstructed at the previous frame.
The flow is then integrated from $t_0$ to $1$.
Unlike Lift4D, which samples independent noise for each frame, we reuse the same noise $\mathbf{Z}^{0}_{0}$ throughout the sequence.
We find that this improves temporal consistency, substantially reducing flips of the canonical space (in which the canonical shape and the pose rotate together, \eg, by 180$^\circ$) and shape variations between consecutive frames (see \cref{sec:supp_same_noise}).

\paragraph{Temporal point correspondences.}
Although the resulting surfaces $\{\mathcal{S}_n\}_{n=0}^{F-1}$ are more consistent, they do not provide explicit correspondences between surface points across frames.
To establish these correspondences, we sample $P$ points $\{\mathbf{p}_i\}_{i=1}^{P}$ on the first reconstructed surface $\mathcal{S}_0$ and fit a time-dependent deformation field $\mathcal{D}$.

For each frame $n$, the field maps every reference point $\mathbf{p}_i$ to its deformed canonical position,
\begin{equation}
  \mathbf{x}^{\mathrm{c}}_{n,i} = \mathbf{p}_i + \mathcal{D}(\mathbf{p}_i, n).
\end{equation}
We optimize $\mathcal{D}$ such that the deformed reference surface matches the reconstructed surface $\mathcal{S}_n$ in canonical space, using a Chamfer-distance objective with deformation-magnitude and temporal regularization.
Since the same reference points are propagated through all frames, their trajectories $\{\mathbf{x}^{\mathrm{c}}_{n,i}\}_{n=0}^{F-1}$ define corresponding point tracks.

Our deformation model follows Lift4D~\cite{litman2026lift4d}, but operates directly on surface points rather than Gaussian splats, without control nodes or photometric supervision.

\paragraph{World-space tracks.}
The canonical tracks capture changes in the object's shape but exclude its global rigid motion.
To recover the latter, we transform the object poses predicted by SAM 3D into world coordinates using camera poses estimated by Depth Anything 3 (DA3)~\cite{da3}, and smooth the resulting object poses over time.

Denoting the smoothed object-to-world scale, rotation, and translation at frame $n$ by $s_n$, $\mathbf{R}_n$, and $\mathbf{t}_n$, the corresponding world-space tracks are
\begin{equation}
  \mathbf{x}^{\mathrm{w}}_{n,i} = s_n \mathbf{R}_n \mathbf{x}^{\mathrm{c}}_{n,i} + \mathbf{t}_n.
\end{equation}

We apply this pipeline to a filtered version of AiM~\cite{zhao2025animalvideo} (see \cref{sec:supp_data}) to construct AiM-Track, containing 7,130 training sequences (528k frames, 23 animal categories) and 233 test sequences from AiM's benchmark.

\subsection{Sampling from Synthetic Animations}
\label{sec:data_synthetic}
We complement our real-world data with 1,653 quadruped animations from Deforming\-Things4D (DT4D)~\cite{dt4d}.
We obtain temporally consistent tracks by sampling points on the mesh faces and keeping their barycentric coordinates fixed across frames.
To obtain their canonical representations, we align each frame to the first context frame using a translation and a rotation about the vertical axis.
We hold out 167 animations for testing.
Further details are provided in \cref{sec:supp_data}.
\section{Method}
\label{sec:method}

A deformable object moves in two ways at once: it changes its pose by rigid motion and its shape by non-rigid deformation.
TrackCast makes this structure explicit by expressing every track in two spaces, a canonical space that factors out the object's pose and holds only its deformation, and a world space that also holds its rigid motion (\cref{fig:method_overview}).
Each track becomes a single token that carries its context and future in both spaces, and a diffusion transformer denoises all tokens jointly by flow matching.

\subsection{Problem Formulation}
\label{sec:setup}
Let $\mathbf{X}_n \in \mathbb{R}^{M \times 3}$ be the positions of $M$ full-surface tracks at frame $n$.
We formulate forecasting as conditional generation, modeling the distribution of the tracks over the next $T$ frames given the tracks of $N$ context frames,
\begin{equation}
  p(\mathbf{X}_{N+1:N+T} \mid \mathbf{X}_{1:N}),
  \label{eq:forecast}
\end{equation}
from which we sample future motions.
This distribution is conditioned on the context tracks alone, in contrast to forecasters that require action or language conditioning.

\subsection{Disentangled Track Diffusion}
\label{sec:track_diffusion}
\begin{figure*}[t]
  \centering
  \includegraphics[width=\textwidth]{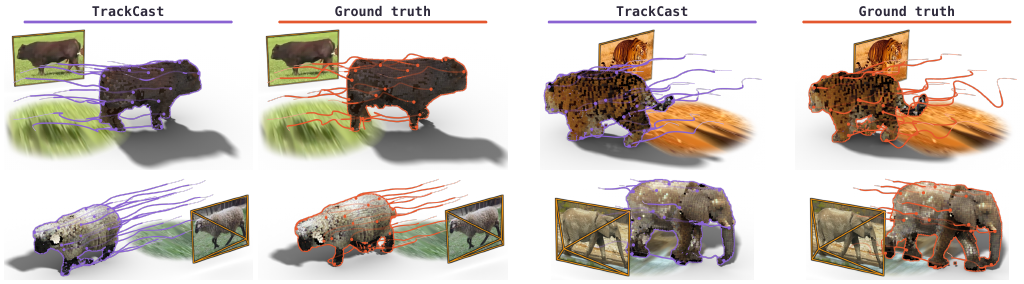}
  \caption{\textbf{Forecasts on held-out AiM-Track videos.} Predictions are in 3D and shown from a novel view, with the input image as a floating card. We show one unconditional sample per video, and points are colored with RGB from SAM 3D's reconstruction of the first context frame.}
  \label{fig:results_gallery}
  \vspace{-8pt}
\end{figure*}

We represent the tracks $\mathbf{X}_n$ in the two spaces of \cref{sec:data_processing}, as $\mathbf{X}^{\mathrm{c}}_n$ in canonical space and $\mathbf{X}^{\mathrm{w}}_n$ in world space.
This suits full-surface tracks, for which the center of the whole object gives the canonical space a natural origin.

\paragraph{Track tokens.}
As in previous 2D track forecasters~\cite{track2act,fmitw}, each token holds one whole track, so the number of tokens is $M$ regardless of the number of frames.
This token efficiency is especially beneficial for full-surface forecasting, which needs more tracks to cover the unseen surface as well as the visible one.
To hold the whole track, each token concatenates a clean context part and a noisy future part, each in both spaces (\cref{fig:method_overview}, left).
The context part holds the track's position at the last context frame and the $N-1$ velocities leading up to it, and the future part its displacements from that position at the $T$ future frames.
A single linear layer maps this concatenation to the transformer's width, and a separate linear layer per space maps each track's final output to its clean future displacements in that space.
Since both spaces share one token, the canonical space adds neither tokens nor attention cost.

\paragraph{Joint flow matching.}
We denote the future displacements of all tracks in both spaces as $\mathbf{Y} = (\mathbf{Y}^{\mathrm{c}}, \mathbf{Y}^{\mathrm{w}})$.
To generate these displacements, we use a diffusion transformer (DiT)~\cite{diffusion_transformer} that processes the $M$ track tokens with full self-attention.

During training, we interpolate between Gaussian noise $\boldsymbol{\epsilon}\sim\mathcal{N}(\mathbf{0},\mathbf{I})$ and the ground-truth displacements $\mathbf{Y}$,
\begin{equation}
  \mathbf{Y}_t = (1-t)\boldsymbol{\epsilon} + t\mathbf{Y},
\end{equation}
where $t\in[0,1]$ denotes the flow time.
We use the same $t$ for both canonical and world displacements, allowing the model to generate deformation and rigid motion jointly.

Following JiT~\cite{Li_2026_CVPR}, the DiT predicts the clean displacements $\hat{\mathbf{Y}}$.
The flow time $t$ is incorporated through adaptive layer normalization (AdaLN), and the model is trained with the flow-matching objective~\cite{flow_matching}
\begin{equation}
  \mathcal{L} = \mathbb{E}_{t,\boldsymbol{\epsilon}} \Big\| \frac{\hat{\mathbf{Y}} - \mathbf{Y}_t}{1-t} - (\mathbf{Y} - \boldsymbol{\epsilon}) \Big\|^2 .
  \label{eq:flow_loss}
\end{equation}
At inference, we integrate the predicted velocity field from Gaussian noise using Euler steps.
The resulting displacements are added to the last context positions, yielding a sample from the conditional distribution in \cref{eq:forecast} in both spaces (\cref{fig:method_overview}, right).

\paragraph{Motion conditioning.}
\label{sec:conditioning}
Our dual-space representation also enables separate conditioning on the object's world-space motion and its deformation through two optional conditioning magnitudes:
\begin{equation}
  \begin{aligned}
    m^{\mathrm{w}} &= \tfrac{1}{M} \textstyle\sum_{i} \big\| \mathbf{x}^{\mathrm{w}}_{N+T,i} - \mathbf{x}^{\mathrm{w}}_{N,i} \big\|, \\
    m^{\mathrm{c}} &= \tfrac{1}{MT} \textstyle\sum_{n=N+1}^{N+T}\sum_i \big\| \mathbf{x}^{\mathrm{c}}_{n,i} - \mathbf{x}^{\mathrm{c}}_{n-1,i} \big\|,
  \end{aligned}
  \label{eq:magnitudes}
\end{equation}
where $\mathbf{x}^{\mathrm{w}}_{n,i}$ and $\mathbf{x}^{\mathrm{c}}_{n,i}$ are the positions of track $i$ in $\mathbf{X}^{\mathrm{w}}_n$ and $\mathbf{X}^{\mathrm{c}}_n$.
The world magnitude $m^{\mathrm{w}}$ measures the average displacement of the tracks in world space from the last context frame to the last future frame, while the canonical magnitude $m^{\mathrm{c}}$ measures their average per-frame deformation in canonical space.
During training, both magnitudes are computed from the ground-truth future and independently dropped for half of the samples.
Following Thakkar et al.~\cite{fmitw}, each magnitude is linearly embedded into the AdaLN conditioning and zeroed when dropped.
Unlike their single 2D displacement, our two magnitudes also provide explicit conditioning on deformation.

\begin{figure*}[t]
  \centering
  \includegraphics[width=\textwidth]{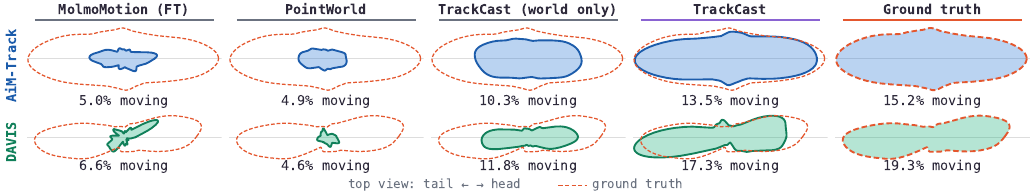}
  \caption{\textbf{Directions of track motion relative to the body.} Each plot shows, seen from above, the directions in which the moving tracks of a forecast (same criterion as in \cref{sec:experimental_setup}) move after removing the object's rigid motion, for AiM-Track \textit{(top)} and DAVIS \textit{(bottom)}; its size scales with the share of tracks that move, and the dashed outline repeats the ground truth \textit{(right)}. One sample per window. TrackCast's tracks follow the ground-truth directions, and nearly as many of them move.}
  \label{fig:leg_directions}
\end{figure*}

\newcommand{\worldonlysinglerow}{\oracle{\mname{TrackCast (world only, 1 sample)}} & \oracle{5.16} & \oracle{6.05} & \oracle{6.23} & \oracle{14.96} & \oracle{9.77} & \oracle{22.91} \\}
\begin{table*}[t]
  \centering
  \caption{\textbf{3D track forecasting error on three datasets.} Canonical EPE measures deformation on the moving tracks (see \cref{sec:experimental_setup}). TrackCast has the lowest canonical EPE on all datasets and, unlike all baselines, is better than Persistence in this metric. Gray: our world-only ablation (\cref{sec:ablation_frames}), not ranked. EPE in voxels; bold: best, underlined: second; --: requires RGB input.}
  \label{tab:main_results}
  \small
  \begin{tabularx}{\textwidth}{l YYY YYY}
    \toprule
    \multirow{2}{*}{Method} & \multicolumn{3}{c}{Canonical EPE$\downarrow$} & \multicolumn{3}{c}{World EPE$\downarrow$} \\
    \cmidrule(lr){2-4} \cmidrule(lr){5-7}
    & AiM-Track & DT4D & DAVIS & AiM-Track & DT4D & DAVIS \\
    \midrule
    \mname{Persistence} & 5.05 & 7.03 & 5.95 & 20.03 & 36.30 & 22.61 \\
    \mname{Constant Velocity} & 8.36 & 14.08 & 10.72 & 23.60 & 37.21 & 27.57 \\
    \mname{Rigid Constant Velocity} & 5.05 & 7.03 & 5.95 & 22.75 & 30.24 & 25.25 \\
    \arrayrulecolor{black!25}\specialrule{0.3pt}{2pt}{2pt}\arrayrulecolor{black}
    \mname{MolmoMotion (zero-shot)}~\cite{zhang2026molmomotion} & 6.66 & -- & 8.77 & 24.34 & -- & 49.73 \\
    \mname{MolmoMotion (fine-tuned)}~\cite{zhang2026molmomotion} & 5.38 & -- & 6.42 & 14.88 & -- & 22.29 \\
    \mname{PointWorld}~\cite{point_world} & 5.30 & 8.37 & 6.43 & \second{14.62} & 14.01 & 18.67 \\
    \arrayrulecolor{black!25}\specialrule{0.3pt}{2pt}{2pt}\arrayrulecolor{black}
    \worldonlysinglerow
    \oracle{\mname{TrackCast (world only, best of 5)}} & \oracle{4.76} & \oracle{5.49} & \oracle{5.63} & \oracle{14.16} & \oracle{9.12} & \oracle{22.26} \\
    \rowcolor{ourrow} \oursbar\mname{\textbf{TrackCast (1 sample)}} & \second{4.74} & \second{3.40} & \second{5.34} & 14.76 & \second{8.96} & \second{17.94} \\
    \rowcolor{ourrow} \oursbar\mname{\textbf{TrackCast (best of 5)}} & \best{4.56} & \best{3.20} & \best{5.11} & \best{14.11} & \best{8.34} & \best{17.20} \\
    \bottomrule
  \end{tabularx}
  \vspace{-8pt}
\end{table*}

\subsection{Implementation Details}
\label{sec:implementation}
We use $M = 2048$ tracks, with $N = 4$ context and $T = 12$ future frames spaced 0.2\,s apart, so 0.6\,s of context gives a 2.4\,s forecast.
We sample training clips from the AiM-Track and DT4D training splits with probabilities 0.8 and 0.2.
The DiT has 380M parameters and is trained for 400k steps with a batch size of 128, which takes about 6 days on 8 AMD Instinct MI210 GPUs.
At inference, sampling uses 16 Euler steps; further details are in \cref{sec:supp_implementation}.

\section{Experiments}
\label{sec:experiments}

We evaluate TrackCast's forecasting accuracy on real-world and synthetic animal datasets, show how motion conditioning steers its forecasts, demonstrate its applicability to robotic manipulation, and ablate its canonical space.

\subsection{Datasets and Metrics}
\label{sec:experimental_setup}

\paragraph{Datasets.}
We evaluate on three test sets: 429 windows from the 146 AiM-Track test sequences long enough for one window, 300 windows from held-out DT4D animations, and 24 windows of 8 quadrupeds from 7 DAVIS~\cite{pont2017davis} sequences selected by clip length and track quality (see \cref{sec:supp_data}).
Each window contains $N$ context frames and $T$ future frames.
To make the test setup fully causal, we construct the AiM-Track and DAVIS input tracks by running the full extraction pipeline (\cref{sec:data_real}) only on the video covered by the context, without access to future frames.
Ground-truth tracks are instead extracted from the complete video.
As the input and ground-truth tracks differ, we align their context trajectories with a similarity transform and transfer the predicted trajectories to the ground-truth tracks by inverse-distance weighting over the four nearest input tracks.

\paragraph{Metrics.}
We report the end-point error (EPE) in world and in canonical space,
\begin{equation}
  \begin{aligned}
    \mathrm{EPE}^{\mathrm{w}} &= \tfrac{1}{MT} \textstyle\sum_{n,i} \big\| \hat{\mathbf{x}}^{\mathrm{w}}_{n,i} - \mathbf{x}^{\mathrm{w}}_{n,i} \big\|, \\
    \mathrm{EPE}^{\mathrm{c}} &= \tfrac{1}{|\mathcal{K}|T} \textstyle\sum_{n,\, i \in \mathcal{K}} \big\| \hat{\mathcal{A}}_n\big(\hat{\mathbf{x}}^{\mathrm{w}}_{n,i}\big) - \mathcal{A}_n\big(\mathbf{x}^{\mathrm{w}}_{n,i}\big) \big\|,
  \end{aligned}
  \label{eq:epe}
\end{equation}
where $\hat{\mathbf{x}}_{n,i}$ and $\mathbf{x}_{n,i}$ are the predicted and ground-truth positions of track $i$ at future frame $n$, and $\hat{\mathcal{A}}_n$ and $\mathcal{A}_n$ are the least-squares similarity transforms that align the predicted and the ground-truth frame $n$, respectively, to the canonical tracks of the last context frame.
Since most of the body hardly deforms, the canonical EPE is averaged over the set $\mathcal{K}$ of moving tracks: after this alignment, a ground-truth track counts as moving if it moves at least twice as far as the median track and at least one voxel.
Canonical EPE thus measures only deformation, and since it is computed from the world prediction, it applies to every method.
EPE is in voxels of SAM 3D's normalized $64^3$ object grid.
Following Thakkar et al.~\cite{fmitw}, we report the best of five samples for TrackCast, chosen per window and metric.
For a fairer comparison with deterministic forecasters, we also report the error of a single sample.

\begin{figure*}[t]
  \centering
  \includegraphics[width=\textwidth]{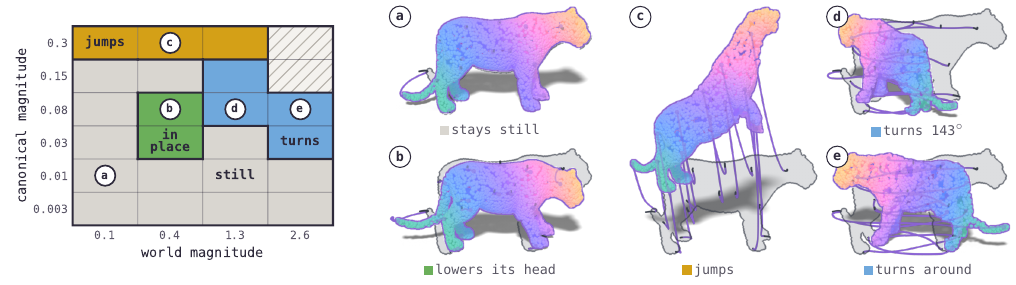}
  \caption{\textbf{Behavior control.} Forecasts for a DT4D puma test clip under different world and canonical conditioning magnitudes (\cref{sec:conditioning}). \textit{(Left)} The behavior at each tested pair of magnitudes, grouped by k-means clustering on motion features (see \cref{sec:supp_implementation}). \textit{(Right)} Forecasts at the marked pairs, with the last context pose in gray. Each forecast is the mean of four samples at the last future frame; (c) is shown mid-jump at the fourth future frame.}
  \label{fig:control_behaviour}
  \vspace{-6pt}
\end{figure*}

\begin{figure*}[t]
  \centering
  \includegraphics{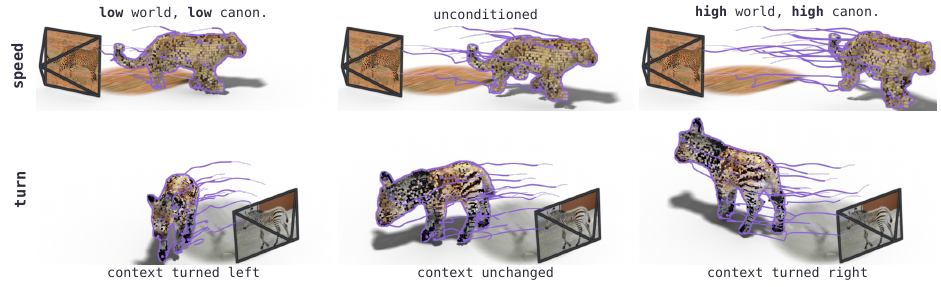}
  \caption{\textbf{Locomotion control.} \textit{(Top)} Setting both conditioning magnitudes low or high (\cref{sec:conditioning}) slows the object down or speeds it up. \textit{(Bottom)} Without conditioning, we gradually rotate the observed 3D context about the vertical axis, by 21$^\circ$ to the left or 15$^\circ$ to the right, and the forecast continues the turn by a further 39$^\circ$ and 29$^\circ$, respectively.}
  \label{fig:control_gallery}
  \vspace{-8pt}
\end{figure*}

\subsection{Baselines}
\label{sec:baselines}
We compare TrackCast, evaluated without motion conditioning, against non-learned and learned baselines.

\paragraph{Non-learned baselines.}
Following Thakkar et al.~\cite{fmitw}, \emph{Persistence} keeps all tracks fixed at their last observed positions, while \emph{Constant Velocity} extrapolates each track using its mean velocity over the context frames; both operate in world space.
\emph{Rigid Constant Velocity} instead extrapolates the object's translation and rotation at their mean context velocities, preserving its last observed shape, so its canonical EPE equals that of Persistence.%
{\widowpenalty=10000\par}

\paragraph{PointWorld.}
PointWorld~\cite{point_world} regresses per-point 3D displacements from scene geometry and robot actions represented as 3D point flows.
We adapt it by replacing the robot actions with the observed context tracks and train it on the same data as TrackCast.
We use its default base configuration (410M parameters vs.~380M for TrackCast).

\paragraph{MolmoMotion.}
MolmoMotion~\cite{zhang2026molmomotion} forecasts 3D point trajectories from RGB frames and a language description of the action, which we derive from the observed context motion.
We evaluate the released model, which decodes greedily, both zero-shot and fine-tuned on the AiM-Track training split.
Since DT4D does not provide RGB frames, we evaluate it only on AiM-Track and DAVIS.
Due to its context-length limit of 8 points per call, we predict 32 tracks per window and interpolate the rest (see \cref{sec:supp_baselines}).

\subsection{Forecast Quality}
\label{sec:results_forecasting}
\begin{figure*}[t]
  \centering
  \includegraphics[width=\textwidth]{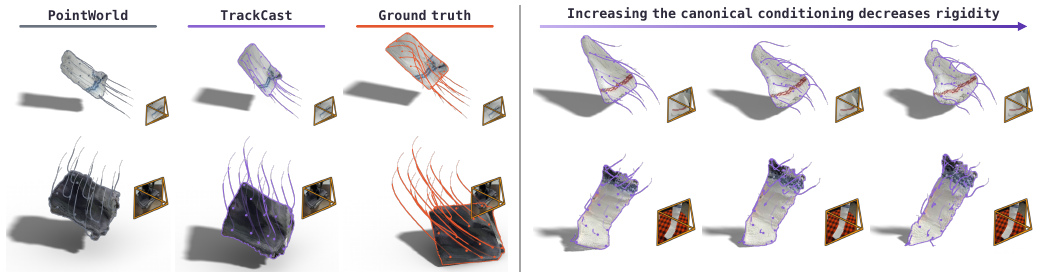}
  \caption{\textbf{Forecasts on held-out DROID videos.} \textit{(Left)} TrackCast forecasts the object's motion closer to the ground truth than PointWorld, both without action conditioning. \textit{(Right)} From the same context, a higher canonical conditioning (\cref{sec:conditioning}) makes the forecast less rigid.}
  \label{fig:robotics_gallery}
  \vspace{-8pt}
\end{figure*}
\Cref{fig:results_gallery} illustrates forecasts of TrackCast on held-out AiM-Track videos.
The predicted motions follow the ground truth in direction and cover similar distances.
Quantitatively, \cref{tab:main_results} shows that TrackCast has a lower canonical EPE than every baseline on all three datasets, both with a single sample and with the best of five.
In this metric, the learned baselines are less accurate than Persistence, whereas TrackCast improves on it, most clearly on DT4D and DAVIS.
Persistence is hard to improve on because the motion of deformable parts, such as the phase of the legs, is difficult to forecast precisely, so repeating the last observed shape already yields a low error.
TrackCast lowers the canonical error further by forecasting how the body deforms (\cref{fig:leg_directions}): its tracks follow the directions of the ground truth, mostly along the head--tail axis, and nearly as many of them move relative to the body, whereas MolmoMotion and PointWorld move only a quarter to a third as many.

With the best of five samples, TrackCast also has the lowest world EPE on all three datasets; a single sample is on par with PointWorld on AiM-Track.
Since world EPE mostly reflects errors in the object's overall position, deformation differences are more apparent in canonical EPE.
These results suggest that explicitly modeling deformation helps TrackCast avoid the overly rigid forecasts of existing 3D forecasters (\cref{sec:ablation_frames}).

\subsection{Behavior and Locomotion Control}
\label{sec:results_controllability}

We control the forecasts in two complementary ways: motion conditioning influences behavior and speed, while modifying the observed context steers their direction.

\paragraph{Behavior control.}
Different behaviors exhibit distinct combinations of world and canonical motion.
By conditioning TrackCast on the two motion magnitudes (\cref{sec:conditioning}), we can encourage different behaviors.
\Cref{fig:control_behaviour} illustrates this on a DT4D puma test clip with a nearly static context: with both magnitudes low, the puma stays still, and a high canonical magnitude makes it jump; in between, varying the world magnitude produces movement in place or turning.
The same control transfers to an Objaverse tiger that is not in the training data (\cref{fig:supp_behavior_other}).%

\paragraph{Locomotion control.}
Setting both magnitudes low or high slows the animal down or speeds it up (\cref{fig:control_gallery}, top).
In \cref{fig:control_gallery} (bottom), we instead steer its direction by gradually rotating the 3D context about the vertical axis, and TrackCast's forecast keeps turning instead of continuing straight.

\subsection{Robotic Manipulation}
\label{sec:results_robotics}

\begin{table}[t]
  \centering
  \caption{\textbf{Forecasting error on robotic manipulation (DROID).} TrackCast (best of 5) has the lowest canonical and world EPE. EPE in voxels; bold: best, underlined: second.}
  \label{tab:robotics}
  \small
  \setlength{\tabcolsep}{2.5pt}
  \begin{tabularx}{\columnwidth}{>{\hspace{2pt}}l >{\hsize=1.15\hsize}Y >{\hsize=0.85\hsize}Y}
    \toprule
    Method & \mbox{Canonical EPE$\downarrow$} & \mbox{World EPE$\downarrow$} \\
    \midrule
    \mname{Persistence} & \second{2.78} & 19.64 \\
    \mname{Constant Velocity} & 3.54 & 18.52 \\
    \mname{Rigid Constant Velocity} & \second{2.78} & 18.28 \\
    \arrayrulecolor{black!25}\specialrule{0.3pt}{2pt}{2pt}\arrayrulecolor{black}
    \mname{PointWorld}~\cite{point_world} & 2.93 & 13.81 \\
    \rowcolor{ourrow} {\setlength{\tabcolsep}{6pt}\oursbar}\mname{\textbf{TrackCast (1 sample)}} & 2.79 & \second{12.81} \\
    \rowcolor{ourrow} {\setlength{\tabcolsep}{6pt}\oursbar}\mname{\textbf{TrackCast (best of 5)}} & \best{2.58} & \best{12.05} \\
    \bottomrule
  \end{tabularx}
  \vspace{-8pt}
\end{table}

To show the versatility of our formulation, we also apply TrackCast to deformable objects in robotic manipulation.

\paragraph{Setup.}
We take clips from DROID~\cite{Khazatsky-RSS-24} tasks such as folding cloth or opening curtains, and extract full-surface tracks of the manipulated object with our pipeline (\cref{sec:data_real}).
We train TrackCast and PointWorld from scratch on 1,564 clips and test on 81 from held-out episodes, with three context and eight future frames.
As for animals, no method is conditioned on the robot's actions.
Additional details are provided in \cref{sec:supp_data}.

\paragraph{Results.}
With the best of five samples, TrackCast has the lowest canonical and world EPE on DROID (\cref{tab:robotics}), improving on both the non-learned baselines and PointWorld.
TrackCast forecasts how the manipulated object moves and deforms (\cref{fig:robotics_gallery}, left).
As on animals, the canonical magnitude controls the deformation: from the same context, a higher magnitude makes the object less rigid (\cref{fig:robotics_gallery}, right).

\subsection{Ablation of the Canonical Space}
\label{sec:ablation_frames}

We ablate the canonical space by training a world-only variant of TrackCast with the same architecture, data, and training schedule, but without canonical inputs or predictions.
This variant uses only the world motion magnitude for conditioning.
Both models are evaluated without conditioning.

The world-only rows of \cref{tab:main_results} show that jointly predicting canonical and world tracks lowers the canonical EPE on all three datasets, and that without the canonical space, the improvement over Persistence shrinks.
Consistent with this, more of the body deforms with the canonical space: the share of tracks that move relative to the body rises to nearly that of the ground truth (\cref{fig:leg_directions}).
The world EPE is lower as well on DT4D and DAVIS, while on AiM-Track both models are on par.

\section{Discussion and Conclusion}
\label{sec:conclusion}

\paragraph{Limitations.}
\label{sec:limitations}
TrackCast forecasts one object at a time.
Extending it to interacting objects and conditioning on the surrounding scene to account for contacts and obstacles are natural future directions.
Moreover, our ground truth is reconstructed rather than directly measured: it depends on SAM 3D, DA3, and our deformation fitting, and its unseen surface is only a plausible completion.
Reconstruction errors can therefore affect both the training data and the ground truth we evaluate against.
Nevertheless, our pipeline requires only videos and can directly benefit from advances in image-to-3D and 4D reconstruction.

\paragraph{Conclusion.}
We introduced the task of full-surface 3D point-track forecasting and an automatic pipeline for constructing AiM-Track from in-the-wild videos.
Our model, TrackCast, jointly forecasts canonical and world-space tracks, capturing both deformation and global motion without requiring a predefined object structure.
On animal motion and robotic manipulation, it forecasts deformation more accurately than state-of-the-art 3D forecasters while matching or improving on all baselines in world error, and on animals, its behavior and speed can be steered through conditioning.
We hope this work contributes toward learning full-surface motion priors from videos at scale, for animation, robotics, and world models.

\section*{Acknowledgments}
\label{sec:acknowledgments}
This work was supported by an EPSRC scholarship (2922572).
We acknowledge EuroHPC Joint Undertaking for awarding us access to LUMI at CSC, Finland (project EHPC-DEV-2026D07-088).
This work was supported in part by Advanced Micro Devices, Inc.\ under the AMD University Program's AI \& HPC Cluster.
We thank Andrea Vedaldi for feedback and advice on the baselines.

\section*{LLM Usage Statement}
\label{sec:llm_usage}
We used Claude Code (Anthropic), an LLM-based coding agent, with Claude 5-family models (mainly Claude Opus 5 to 5.5) throughout this work.
\emph{Ideas:} all core research ideas, the method, and the claims are ours; the agent helped refine them.
\emph{Code, data, and experiments:} following our instructions, the agent wrote and ran the code for the data pipeline, training, and evaluation, and suggested and implemented analyses.
We reviewed the code, the experiment plans, and the results, and every reported number is produced by these scripts.
\emph{Figures:} the agent wrote the scripts that render and arrange the figures from model outputs and dataset frames; no figure content comes from an image-generation model.
\emph{Writing and related work:} the agent drafted text from our notes, suggested edits, searched for related work, and checked every citation against its source.
We edited and approved all text and chose the references.
Generative models that are part of the method or the baselines, such as SAM 3D in our data pipeline (\cref{sec:data_processing}) and MolmoMotion, are described in the paper.
We take full responsibility for all content of this paper, including its text, code, data, results, figures, and claims.

  \clearpage
\SuppFloats\maketitlesupplementary
\setcounter{section}{0}\renewcommand{\thesection}{\Alph{section}}\renewcommand{\theHsection}{supp.\Alph{section}}
\setcounter{figure}{0}\renewcommand{\thefigure}{S\arabic{figure}}\renewcommand{\theHfigure}{supp.\arabic{figure}}
\setcounter{table}{0}\renewcommand{\thetable}{S\arabic{table}}\renewcommand{\theHtable}{supp.\arabic{table}}
\setcounter{equation}{0}\renewcommand{\theequation}{S\arabic{equation}}\renewcommand{\theHequation}{supp.\arabic{equation}}

\renewcommand{\creflastconjunction}{, and\nobreakspace}%
We strongly encourage readers to view the videos on our \href{https://trackcast.jacob-lin.com/}{project page} (\cref{sec:supp_videos}), as forecasts are best judged in motion.
\Cref{sec:supp_behavior_other,sec:supp_significance,sec:supp_dt4d_validation} support claims of the main paper: behavior control on an unseen dataset (\cref{sec:supp_behavior_other}), the significance of the single-sample results (\cref{sec:supp_significance}), and a check that evaluating against extracted instead of true tracks leads to the same conclusions (\cref{sec:supp_dt4d_validation}).
The remaining sections give the implementation, data-pipeline, and baseline details.

\section{Video Results}
\label{sec:supp_videos}
Our project page ({\urlstyle{same}\url{https://trackcast.jacob-lin.com/}}) shows our forecasts in motion and is the most direct way to judge them.

The page has four parts.
The first shows forecasts of TrackCast, PointWorld, and MolmoMotion next to the ground truth for nine held-out sequences from AiM-Track and DAVIS, in a 3D viewer that can be rotated.
The second shows eight AiM-Track training sequences as full-surface point tracks, each with its input video.
The third shows behavior control on an animated tiger from Objaverse as a looping video.
The fourth shows forecasts on held-out DROID videos.

\section{Behavior Control on an Unseen Dataset}
\label{sec:supp_behavior_other}
\begin{figure*}[t]
  \centering
  \includegraphics[width=\textwidth]{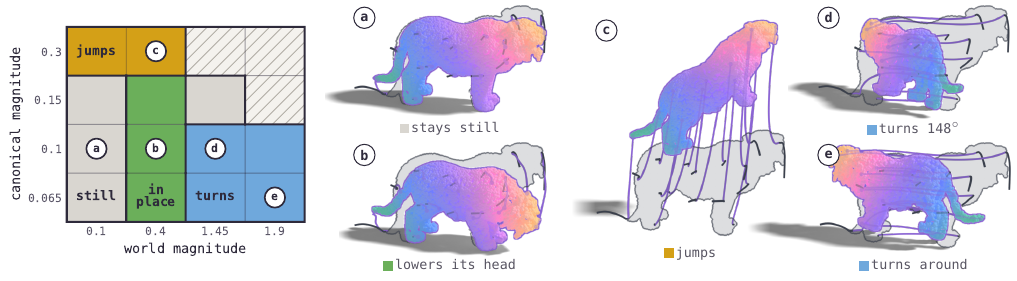}
    \caption{\textbf{Behavior control on an Objaverse tiger.} Its behavior map, computed as in \cref{fig:control_behaviour} (hatched: no consistent behavior), and the forecasts at the marked pairs. Each forecast is the mean of four samples at the last future frame, with the last context pose in gray; the jump is shown mid-jump. Tiger model: ``Tiger attack eat v01'' by shahabahmed, CC BY 4.0.}
  \label{fig:supp_behavior_other}
\end{figure*}
\Cref{fig:supp_behavior_other} shows behavior control (\cref{fig:control_behaviour}) on an animated tiger from Objaverse~\cite{deitke2023objaverse}, a dataset unseen during training.
We prepare its tracks as for the DT4D clips (\cref{sec:supp_data}) and compute its map as in \cref{sec:supp_behavior_map}, on the pairs of magnitudes shown in \cref{fig:supp_behavior_other}.
At the marked pairs, the tiger stays still, lowers its head, jumps, turns, and turns around.

\section{Significance of the Single-Sample Results}
\label{sec:supp_significance}
\Cref{tab:supp_significance} gives the significance of the single-sample results of \cref{tab:main_results}, from a bootstrap that resamples whole test sequences, so that the windows of a sequence stay together.
In canonical EPE, TrackCast's single sample is significantly more accurate in every comparison.
In world EPE, it is on par with PointWorld on AiM-Track and DAVIS and more accurate on DT4D.

\begin{table*}[b]
  \centering
  \caption{\textbf{Significance of the single-sample results.} Each cell gives the mean per-window difference in EPE between TrackCast (1 sample) and a method of \cref{tab:main_results}, with its 95\,\% bootstrap confidence interval over test sequences (10,000 resamples). Negative values mean that TrackCast is more accurate, and bold intervals exclude zero. The world-only variant also uses its first sample. EPE in voxels; --: requires RGB input.}
  \label{tab:supp_significance}
  \small
  \begin{tabular}{@{}l ccc@{}}
    \toprule
    TrackCast (1 sample) minus & AiM-Track & DT4D & DAVIS \\
    \midrule
    \multicolumn{4}{@{}l}{\textit{Canonical EPE}} \\
    \mname{Persistence} & \textbf{$-$0.31 [$-$0.38, $-$0.24]} & \textbf{$-$3.63 [$-$3.94, $-$3.31]} & \textbf{$-$0.61 [$-$1.17, $-$0.09]} \\
    \mname{PointWorld} & \textbf{$-$0.56 [$-$0.68, $-$0.44]} & \textbf{$-$4.96 [$-$5.30, $-$4.62]} & \textbf{$-$1.09 [$-$1.87, $-$0.43]} \\
    \mname{MolmoMotion (fine-tuned)} & \textbf{$-$0.64 [$-$0.77, $-$0.51]} & -- & \textbf{$-$1.07 [$-$1.47, $-$0.66]} \\
    \mname{World-only variant} & \textbf{$-$0.42 [$-$0.50, $-$0.35]} & \textbf{$-$2.64 [$-$2.84, $-$2.44]} & \textbf{$-$0.88 [$-$1.71, $-$0.20]} \\
    \midrule
    \multicolumn{4}{@{}l}{\textit{World EPE}} \\
    \mname{Persistence} & \textbf{$-$5.27 [$-$7.17, $-$3.51]} & \textbf{$-$27.34 [$-$30.95, $-$23.94]} & $-$4.67 [$-$8.90, $+$0.14] \\
    \mname{PointWorld} & $+$0.14 [$-$0.35, $+$0.68] & \textbf{$-$5.05 [$-$6.54, $-$3.64]} & $-$0.73 [$-$3.15, $+$1.98] \\
    \mname{MolmoMotion (fine-tuned)} & $-$0.12 [$-$0.86, $+$0.66] & -- & \textbf{$-$4.35 [$-$6.64, $-$2.04]} \\
    \mname{World-only variant} & $-$0.20 [$-$0.65, $+$0.28] & $-$0.81 [$-$1.57, $+$0.07] & \textbf{$-$4.98 [$-$9.64, $-$1.30]} \\    \bottomrule
  \end{tabular}
\end{table*}

\paragraph{Track groups.}
\Cref{tab:supp_track_groups} splits the canonical EPE on AiM-Track and DAVIS by track group.
On the moving tracks, the single sample improves on Persistence by the same margin whether a track is visible in the first context frame (Seen) or not (Unseen): about 0.3 voxels on AiM-Track and 0.6 on DAVIS.
All four margins are significant under the bootstrap used for \cref{tab:supp_significance}.
On the static tracks, where Persistence is a strong forecast by construction, the single sample stays within 0.07 voxels of it, while PointWorld and MolmoMotion are 0.46 to 0.94 voxels above.

\begin{table}[b]
  \centering
  \caption{\textbf{Canonical EPE by track group.} Canonical EPE of \cref{tab:main_results} on the moving tracks that are visible in the first context frame (Seen) or not (Unseen), and on all other tracks (Static). EPE in voxels; bold: best, underlined: second.}
  \label{tab:supp_track_groups}
  \small
  \setlength{\tabcolsep}{1.5pt}
  \begin{tabularx}{\columnwidth}{>{\hspace{2pt}}l YYY}
    \toprule
    & Seen & Unseen & Static \\
    \midrule
    \multicolumn{4}{@{\hspace{3.5pt}}l}{\textit{AiM-Track}} \\
    \mname{Persistence} & 5.15 & 4.96 & \best{2.11} \\
    \mname{PointWorld} & 5.39 & 5.22 & 2.57 \\
    \mname{MolmoMotion (fine-tuned)} & 5.48 & 5.29 & 2.60 \\
    \rowcolor{ourrow} {\setlength{\tabcolsep}{6pt}\oursbar}\mname{\textbf{TrackCast (1 sample)}} & \second{4.85} & \second{4.66} & 2.18 \\
    \rowcolor{ourrow} {\setlength{\tabcolsep}{6pt}\oursbar}\mname{\textbf{TrackCast (best of 5)}} & \best{4.64} & \best{4.47} & \second{2.12} \\
    \midrule
    \multicolumn{4}{@{\hspace{3.5pt}}l}{\textit{DAVIS}} \\
    \mname{Persistence} & 6.11 & 5.85 & \second{2.40} \\
    \mname{PointWorld} & 6.64 & 6.33 & 3.34 \\
    \mname{MolmoMotion (fine-tuned)} & 6.89 & 6.31 & 2.93 \\
    \rowcolor{ourrow} {\setlength{\tabcolsep}{6pt}\oursbar}\mname{\textbf{TrackCast (1 sample)}} & \second{5.48} & \second{5.25} & 2.43 \\
    \rowcolor{ourrow} {\setlength{\tabcolsep}{6pt}\oursbar}\mname{\textbf{TrackCast (best of 5)}} & \best{5.25} & \best{5.00} & \best{2.36} \\
    \bottomrule
  \end{tabularx}
\end{table}

\section{Validating Evaluation Against Extracted Tracks}
\label{sec:supp_dt4d_validation}
\begin{table}[t]
  \centering
  \caption{\textbf{Scoring against extracted tracks on rendered DT4D.} Forecasting error on the 300 rendered DT4D test windows from context extracted by our pipeline, scored against the true tracks or against the tracks extracted from the whole video (extr.), as AiM-Track and DAVIS are scored in \cref{tab:main_results}. TrackCast has the lowest canonical EPE against both. EPE in voxels; bold: best, underlined: second.}
  \label{tab:supp_dt4d_validation}
  \small
  \setlength{\tabcolsep}{1.5pt}
  \begin{tabularx}{\columnwidth}{>{\hspace{2pt}}l YY YY}
    \toprule
    & \multicolumn{2}{c}{Canonical EPE$\downarrow$} & \multicolumn{2}{c}{World EPE$\downarrow$} \\
    \cmidrule(lr){2-3} \cmidrule(lr){4-5}
    Scored against & true & extr. & true & extr. \\
    \midrule
    \mname{Persistence} & 7.01 & 6.29 & 37.60 & 33.59 \\
    \mname{Constant Velocity} & 11.02 & 10.85 & 40.53 & 45.80 \\
    \mname{PointWorld}~\cite{point_world} & 7.85 & 7.19 & \second{26.88} & 26.89 \\
    \rowcolor{ourrow} {\setlength{\tabcolsep}{6pt}\oursbar}\mname{\textbf{TrackCast (1 sample)}} & \second{6.73} & \second{5.77} & 27.06 & \second{26.60} \\
    \rowcolor{ourrow} {\setlength{\tabcolsep}{6pt}\oursbar}\mname{\textbf{TrackCast (best of 5)}} & \best{6.51} & \best{5.54} & \best{26.35} & \best{25.91} \\
    \bottomrule
  \end{tabularx}
\end{table}

In \cref{tab:main_results}, the forecasts on AiM-Track and DAVIS are scored against tracks extracted by our pipeline, as in-the-wild videos have no 3D ground truth.
To test whether scoring against extracted instead of true tracks changes the comparison between methods, we apply the same protocol to the 300 DT4D test windows of \cref{tab:main_results}, whose true tracks are known.
We render each window in Blender as a video of 31 frames at 10\,fps, with a procedural fur texture, a textured ground, and a camera that moves like the estimated camera of an AiM-Track sequence and follows the animal.
We then run the unchanged pipeline (\cref{sec:data_real}) on the frames, cropped around the rendered masks, twice: on the whole video, and only on the context frames, as for the test inputs of \cref{tab:main_results}.
Finally, the methods of \cref{tab:supp_dt4d_validation} forecast each window from the context-only extraction, and we score the forecasts as in \cref{sec:experimental_setup}, both against the true tracks and against the tracks extracted from the whole video.

Against both, TrackCast has a lower canonical EPE than all baselines, and the margin of its single sample over Persistence and PointWorld is significant (95\,\% bootstrap confidence intervals over windows).
In world EPE, it is on par with PointWorld (\cref{tab:supp_dt4d_validation}).
Scoring against extracted instead of true tracks therefore leads to the same conclusions.

\section{Additional Implementation Details}
\label{sec:supp_implementation}
\begin{table}[b]
  \centering
  \caption{\textbf{Additional implementation details.} Hyperparameters of TrackCast that are not given in the main paper.}
  \label{tab:supp_implementation}
  \small
  \begin{tabularx}{\columnwidth}{@{}l >{\raggedright\arraybackslash}X@{}}
    \toprule
    \multicolumn{2}{@{}l}{\textit{Model}} \\
    Backbone & 20 blocks, width 1024, 16 heads \\
    Attention & QK normalization \\
    Conditioning & includes the frame times (Fourier features) \\
    Output & a zero-initialized head per space \\
    Frame spacing & 0.2\,s, every second frame of the 10\,fps tracks \\
    \midrule
    \multicolumn{2}{@{}l}{\textit{Flow matching}} \\
    Velocity & $1-t$ clamped at 0.05 in \cref{eq:flow_loss}~\cite{Li_2026_CVPR} \\
    Flow time & logit-normal ($\mu = -0.8$, $\sigma = 0.8$)~\cite{Li_2026_CVPR} \\
    \midrule
    \multicolumn{2}{@{}l}{\textit{Training}} \\
    Optimizer & AdamW, $\beta = (0.9, 0.95)$, weight decay 0.01 \\
    Learning rate & $10^{-4}$, 2k warm-up steps, cosine to $10^{-5}$ \\
    Gradient clipping & 5 \\
    Precision & bfloat16 \\
    EMA & decay 0.9999, used for evaluation \\
    \midrule
    \multicolumn{2}{@{}l}{\textit{Inference}} \\
    Single sample & the first of the five samples \\
    DAVIS (24\,fps) & resampled to 10\,fps \\
    \bottomrule
  \end{tabularx}
\end{table}

\paragraph{World space.}
For each window, we express the world tracks in the camera space of its first context frame, after removing the camera motion with DA3's camera poses, and in units of the object's scale at that frame.
At the first context frame, the world tracks therefore differ from the canonical tracks only by the object's rotation and translation relative to the camera.

\paragraph{Track tokens.}
For each space $\mathrm{s} \in \{\mathrm{c}, \mathrm{w}\}$, the token of track $i$ holds a clean context part $\mathbf{h}^{\mathrm{s}}_i$ and a future part $\mathbf{y}^{\mathrm{s}}_i$, which enters in its noisy form (\cref{sec:track_diffusion}),
\begin{equation}
  \begin{aligned}
    \mathbf{h}^{\mathrm{s}}_i &= \big[\, \mathbf{x}^{\mathrm{s}}_{N,i},\; \gamma^{\mathrm{s}} \big( \mathbf{x}^{\mathrm{s}}_{n,i} - \mathbf{x}^{\mathrm{s}}_{n-1,i} \big)_{n=2}^{N} \,\big], \\
    \mathbf{y}^{\mathrm{s}}_i &= \gamma^{\mathrm{s}} \big( \mathbf{x}^{\mathrm{s}}_{n,i} - \mathbf{x}^{\mathrm{s}}_{N,i} \big)_{n=N+1}^{N+T}.
  \end{aligned}
  \label{eq:supp_token}
\end{equation}
With $N = 4$ and $T = 12$, the two context parts ($2 \times 12$ values) and the two future parts ($2 \times 36$ values) give 96 values per token.
The factors $\gamma^{\mathrm{c}} = 15$ and $\gamma^{\mathrm{w}} = 3$ bring the velocities and displacements of both spaces to similar magnitudes, and the predicted displacements are divided by the same factors.

\paragraph{Conditioning magnitudes.}
The conditioning magnitudes (\cref{eq:magnitudes}) enter their linear embeddings as $\log(1 + m^{\mathrm{w}})$ and $\log(1 + T m^{\mathrm{c}})$.
As both are computed from the future, for \cref{fig:control_gallery} (top), we extrapolate them from the context,
\begin{equation}
  \begin{aligned}
    \tilde{m}^{\mathrm{w}} &= \frac{T}{N-1} \cdot \frac{1}{M} \sum_{i} \big\| \mathbf{x}^{\mathrm{w}}_{N,i} - \mathbf{x}^{\mathrm{w}}_{1,i} \big\|, \\
        \tilde{m}^{\mathrm{c}} &= \frac{1}{M(N-1)} \sum_{n=2}^{N}\sum_i \big\| \mathbf{x}^{\mathrm{c}}_{n,i} - \mathbf{x}^{\mathrm{c}}_{n-1,i} \big\|,
  \end{aligned}
  \label{eq:supp_extrap}
\end{equation}
and multiply both by 0.25 for the low and by 2 for the high setting.
The factor $T/(N-1) = 4$ accounts for the forecast spanning four times as long as the context; $\tilde{m}^{\mathrm{c}}$ is a distance per frame and needs no such factor.

\paragraph{Context rotation.}
To steer the direction of the forecast in \cref{fig:control_gallery} (bottom), we rotate the world-space context progressively about the object's vertical axis, as if the object had started to turn.
Context frame $n = 1, \dots, N$ is rotated by $(n-1)\,\theta$ about its centroid, and each step of the centroid path is rotated by the mean angle of its two frames.
The first context frame is thus unchanged, and the last is turned by $(N-1)\,\theta$ (21$^\circ$ to the left and 15$^\circ$ to the right in the figure).
The vertical axis is the object's canonical up axis mapped into world space by the pose of each context frame and averaged over these frames.
The canonical context is not modified.

\paragraph{Tracks.}
The tables use a random subset of 2048 tracks per window, as in training.
For renders with more tracks, the model forecasts a random subset of 2048, and every other track $j$ follows the displacement of its four nearest forecast tracks $\mathcal{N}_j$, weighted by inverse distance,
\begin{equation}
  \hat{\mathbf{x}}_{n,j} = \mathbf{x}_{N,j} + \frac{\sum_{i \in \mathcal{N}_j} d_{ij}^{-1}\, (\hat{\mathbf{x}}_{n,i} - \mathbf{x}_{N,i})}{\sum_{i \in \mathcal{N}_j} d_{ij}^{-1}},
  \label{eq:supp_idw}
\end{equation}
where $\hat{\mathbf{x}}_{n,i}$ is the forecast position of track $i$ at future frame $n$, and $d_{ij}$ is the distance between tracks $i$ and $j$ in canonical space at the last context frame.

\paragraph{Canonical and world outputs.}
TrackCast also outputs a canonical forecast, which \cref{tab:main_results} does not score.
It closely agrees with the world forecast after a similarity alignment per frame, and scoring it instead gives the same or a lower canonical EPE on all three test sets (4.72 vs.\ 4.74 on AiM-Track with a single sample).

\paragraph{Moving tracks.}
The motion that decides whether a ground-truth track counts as moving (\cref{sec:experimental_setup}) is its mean distance from its position at the last context frame over the forecast, measured after the alignment to the canonical space.
Seven of the 300 DT4D windows have no moving track and do not enter the canonical EPE.

\paragraph{Test windows.}
From every AiM-Track and DAVIS test sequence that is at least one window long, we take three windows whose start frames are evenly spaced from the first to the last possible start; on short sequences, windows with the same start are counted once.

\paragraph{Behavior map.}
\label{sec:supp_behavior_map}
For the map in \cref{fig:control_behaviour}, we forecast the DT4D puma test clip at the 24 pairs of magnitudes on its axes and average four samples per pair.
From each averaged forecast, we compute five features.
Three describe the motion of the whole body: the highest rise of the centroid above its last context position, the horizontal travel of the centroid at the last future frame, and the turn of the body about the vertical axis at the last future frame.
Two describe the deformation in canonical space: the largest drop of the upper front or back end of the body, and the articulation, \ie, the mean RMS deviation of the tracks from their temporal mean.
Distances are in body radii, the mean distance of the canonical points from their centroid at the last context frame.

We standardize each feature over the 24 pairs and cluster the pairs by k-means with $k = 5$.
Four clusters correspond to the behaviors in the map: still, turns, jumps, and in place.
We hatch the fifth cluster (world 2.6 with canonical 0.15 or 0.3), whose forecasts do not show a consistent behavior.%
{\widowpenalty=10000\par}

\section{Data Pipeline Details}
\label{sec:supp_data}
We first describe how we build AiM-Track with the pipeline of \cref{sec:data_real}, and then the DAVIS test set and the DT4D and DROID clips.

\paragraph{Clip selection.}
Of AiM's 29,749 uncurated clips, \ie, those outside its benchmark, we keep those that pass five quality filters computed from AiM's own masks, occlusion maps, keypoints, and optical flow.
The filters bound the occluded share of the mask contour, the optical-flow magnitude, the keypoint articulation (the motion of the keypoints relative to their mean motion), the number of confidently detected keypoints, and the confidence of the limb keypoints.
We exclude from training every uncurated clip whose source video also contains a benchmark clip, so no test footage is seen in training.

\paragraph{Frames.}
AiM's uncurated clips come without RGB.
We restore their frames from the source YouTube videos as $512^2$ crops, using AiM's own crop boxes; AiM's benchmark clips include their crops.
Frames whose mask has fewer than 500 pixels are dropped.

\paragraph{Depth and cameras.}
DA3 runs jointly over all frames of a clip, with the middle frame as reference, and gives per-frame depth, intrinsics, and camera poses (we use \texttt{DA3NESTED-GIANT-LARGE-1.1}).
Its depth thus has one consistent scale across frames, so the poses need no per-frame rescaling.
As in Lift4D~\cite{litman2026lift4d}, its point maps replace those of MoGe~\cite{wang2025moge} as SAM 3D's depth conditioning.

\paragraph{Reconstruction.}
For the tracks, we run only SAM 3D's sparse-structure stage, which predicts the coarse geometry on a $64^3$ occupancy grid; its second stage only provides the point colors in our figures.
From the grid, we seal small gaps in the surface with a morphological closing, fill the interior, and take the boundary voxels of the resulting solid (6-connectivity) as surface points.
The noise in \cref{eq:causal_init} is fixed per clip, seeded by the clip name, and $t_0 = 0.1$; every later frame is denoised over the remaining schedule in 25 steps.

\paragraph{Silhouette refinement.}
Similar to the mask-based rendering supervision of Lift4D~\cite{litman2026lift4d}, we refine SAM 3D's pose in every frame so that the reconstruction's silhouette matches the mask.
With Adam, we maximize the soft IoU between a differentiable splat of the surface points and the mask, first over scale and translation, then also over rotation.

\paragraph{World space and smoothing.}
We move the refined poses into DA3's world space, \ie, the camera space of the middle frame, with DA3's camera poses.
We then smooth them with a centered moving average over five frames, averaging rotations as quaternions and scales in log space.

\paragraph{Deformation field.}
The deformation field $\mathcal{D}$ of \cref{sec:data_real} is one coordinate MLP per sequence.
It has eight layers of width 256, takes a point and Fourier features of time as input, and has a zero-initialized output layer, so that it starts as the identity.
We fit it by minimizing the sum over the frames $n$ of
\begin{equation}
  \begin{aligned}
    \mathcal{L}_n ={}& \Big(\frac{s_n}{s_0}\Big)^{2}\, d_{\mathrm{CD}}\big( \{\mathbf{x}^{\mathrm{c}}_{n,i}\}_{i=1}^{P},\, \mathcal{S}_n \big) \\
    &+ \frac{\lambda_{1}}{P} \sum_i \big\| \mathcal{D}(\mathbf{p}_i, n) \big\|^2 \\
    &+ \frac{\lambda_{2}}{P} \sum_i \big\| \mathcal{D}(\mathbf{p}_i, n+1) - \mathcal{D}(\mathbf{p}_i, n) \big\|^2,
  \end{aligned}
  \label{eq:supp_deform_loss}
\end{equation}
where $\mathbf{x}^{\mathrm{c}}_{n,i} = \mathbf{p}_i + \mathcal{D}(\mathbf{p}_i, n)$ are the deformed points of the $P$ surface points $\mathbf{p}_i$ of the first frame, $d_{\mathrm{CD}}$ is the squared Chamfer distance, $s_n$ is the smoothed object scale, and $\lambda_1 = 3 \cdot 10^{-4}$ and $\lambda_2 = 0.3$ are fixed weights.
The first term pulls the deformed points onto the frame's surface $\mathcal{S}_n$, with the distance measured at the frame's object scale relative to the first frame, and the other two keep the displacements small and smooth over time.

\paragraph{Flip filtering.}
SAM 3D reconstructs each frame independently, so its output can change abruptly between consecutive frames.
We distinguish three kinds of such discontinuities.
In a \emph{canonical flip}, the canonical shape turns, \eg, by $180^\circ$, and the pose compensates, so the posed shape stays the same.
In a \emph{pose flip}, only the pose turns, so the posed shape suddenly faces another direction.
In a \emph{shape jump}, the canonical shape changes abruptly (the voxel sets of consecutive frames overlap with an IoU below 0.2), while the rotation changes by less than $30^\circ$.
A larger rotation jump is a flip if it is close to one of the 24 rotations that map the canonical cube onto itself.
The flip is canonical if applying this rotation to the canonical shape makes it agree better with the shapes before the jump, and a pose flip otherwise.

We handle the discontinuities in two steps.
First, we undo the canonical flips, so that all frames of a clip share one canonical space.
Second, we split the clip into segments at every flip of either kind, every shape jump, every other rotation jump above $60^\circ$, and where the source video becomes unusable.
Segments whose pose is flipped relative to the clip's first frame are discarded, and every other segment that is long enough becomes a sequence of its own, with its own pose smoothing and deformation fit.

This gives 7,130 training sequences, each at least one window long (5,831 whole clips and 1,299 segments), and 233 test sequences from 226 of AiM's 230 benchmark clips.

\paragraph{DAVIS test set.}
We select the DAVIS test instances by their extracted ground truth alone, without using any forecast.
Among the DAVIS 2017 animal instances, we take every quadruped mammal that meets two criteria.
Its extracted ground truth must span at least 73 frames, so that all 12 future frames of a window lie inside the clip.
Its extracted tracks, projected into the frames, must match its instance mask with a mean IoU of at least 0.7.
This gives eight instances from seven videos (bear, camel, cows, elephant, goat, rhino, and two pigs), with three windows each.
The pipeline uses DAVIS's ground-truth instance masks.

\paragraph{DT4D clips.}
We use 1,653 animations of 48 quadruped characters from DT4D (excluding dragons, dinosaurs, and crocodiles).
We split them randomly by animation into 1,486 for training and 167 for testing.

DT4D's animations are short, so a clip chains animations of the same character whose last and first poses match, \ie, with a rigid alignment residual below 1\,\% of the body diagonal and similar velocities at the join.
The next animation is attached with this rigid transform, so the root motion is kept.
Clips are chained only from animations of their own split, so no test clip shares an animation with a training clip.

The animations run at 25\,fps, so our frames, spaced 0.2\,s apart, are every fifth animation frame.
Each character is scaled so that its longest extent matches the object size of AiM-Track in SAM 3D's grid.
On the first context frame's mesh, we sample 2048 points uniformly by area, new for every drawn clip, so the same animation gives different tracks each time.
The 300 test windows are drawn once in this way from the test animations.
Their context and ground-truth tracks are the same mesh points, so the DT4D results of \cref{tab:main_results} involve neither the extraction pipeline nor a transfer between track sets.

The canonical origin is the bounding-box center of the first context frame, and each frame's alignment to the first context frame (\cref{sec:data_synthetic}) is fitted by least squares over all sampled points.
DT4D has no camera, so we express the world tracks in the camera space of a random AiM-Track training frame, using the transform from canonical space to that camera (one per DT4D clip).

\paragraph{DROID clips.}
We use four DROID task categories that typically involve deformable objects: folding, hanging, cleaning with a cloth, and opening or closing curtains.
No episode appears in more than one split.
The frames are about 0.14\,s apart (every second frame of DROID's video), so the forecast spans about 1.1\,s.

We train TrackCast and PointWorld for 10k steps with a batch size of 128 on the training clips and use the last checkpoint of each.
TrackCast keeps the backbone and flow-matching settings of \cref{tab:supp_implementation} but uses 500 warm-up steps and an EMA decay of 0.999.

\subsection{Same Noise for All Frames}
\label{sec:supp_same_noise}
Here we show that starting every frame of a clip from the same noise (\cref{eq:causal_init}), instead of new noise per frame as in Lift4D~\cite{litman2026lift4d}, improves temporal consistency, as stated in \cref{sec:data_real}.

The discontinuities between SAM 3D's per-frame reconstructions would corrupt the tracks: a canonical flip makes the canonical tracks jump although the animal did not move, and changes between the per-frame shapes enter the fitted deformation.

\paragraph{Comparison.}
We reconstruct AiM's 230 benchmark clips (11,061 frames) once with the same and once with new noise, keeping everything else equal, and measure the reconstructions before the flip filtering cuts the clips, so that every discontinuity counts.
Without 3D ground truth, we measure how consistent the orientation and the shape are between consecutive frames, and how well the reconstructions fit the masks (\cref{tab:supp_same_noise}).

\paragraph{Results.}
With the same noise, the reconstructions are much more consistent over time: rotation jumps are about six times rarer, canonical flips almost disappear, and the shape changes less than half as much between frames.
The deformation field also fits the per-frame surfaces more closely.
These differences are significant (Wilcoxon signed-rank test over clips, $p < 10^{-4}$) and remain on the clips without any rotation jump, so they are not only due to fewer flips.
New noise fits each frame's mask slightly better, as each frame is reconstructed on its own.

\begin{table}[t]
  \centering
  \caption{\textbf{Temporal consistency with the same and with new noise.} SAM 3D reconstructions of AiM's 230 benchmark clips with the same noise for all frames (ours) or new noise per frame (as in Lift4D~\cite{litman2026lift4d}), measured before the flip filtering cuts the clips; rotation steps and Chamfer distances after undoing canonical flips. Means over clips; bold: better.}
  \label{tab:supp_same_noise}
  \small
  \begin{tabularx}{\columnwidth}{@{}X cc@{}}
    \toprule
    & New noise & Same noise \\
    & (per frame) & (ours) \\
    \midrule
    \multicolumn{3}{@{}l}{\textit{Orientation}} \\
    Jumps $> 30^\circ$ per clip & 1.77 & \textbf{0.28} \\
    Clips with a canonical flip (\%) & 41.7 & \textbf{1.7} \\
    Median rotation step ($^\circ$) & 6.40 & \textbf{3.68} \\
    \midrule
    \multicolumn{3}{@{}l}{\textit{Shape}} \\
    Chamfer between frames (voxels) & 0.71 & \textbf{0.29} \\
    \midrule
    \multicolumn{3}{@{}l}{\textit{Fit}} \\
    Silhouette IoU after refinement $\uparrow$ & \textbf{0.835} & 0.827 \\
    Deformation-fit residual ($\times 10^{-3}$) & 3.47 & \textbf{2.41} \\
    \bottomrule
  \end{tabularx}
\end{table}

\section{Baseline Details}
\label{sec:supp_baselines}
\paragraph{PointWorld.}
We train PointWorld for 200k steps with a batch size of 44 and the optimizer settings of its released code, and use the final checkpoint.
Evaluated on the AiM-Track test set every 10k steps, its world EPE levels off after the first 20k steps (between 14.4 and 14.9), and its canonical EPE stays between 5.3 and 5.5, above that of Persistence (5.05) at every checkpoint.

\paragraph{MolmoMotion.}
We use the released model of MolmoMotion~\cite{zhang2026molmomotion}, \texttt{MolmoMotion-4B-H3-F30}.
It decodes autoregressively and greedily, so it makes one deterministic forecast per window; the flow-matching variant of their paper is not released.
The released weights omit the pixel-sampled point feature of their paper, so a query consists of the RGB frames and, as text, the caption and the 3D history.
The model takes three frames, so it receives the last context frame and the two frames before it at the tracks' frame rate of 10\,fps.
The 3D points are in DA3's metric units in the camera space of the last context frame, and the RGB frames are the $512^2$ crops.
We stop generation after 24 frames, which span our forecast horizon, and take every second generated frame as one of our 12 future frames.

At eight points per call (Sec.~6 and App.~E of~\cite{zhang2026molmomotion}), all 2048 tracks would take 256 calls per window, so we forecast 32 tracks in four calls, chosen by farthest-point sampling on the canonical tracks of the last context frame, visible and occluded alike.
Every other track is interpolated from its four nearest forecast tracks by \cref{eq:supp_idw}.
MolmoMotion's error is not caused by this interpolation: on the tracks it forecasts itself, both versions are less accurate than Persistence, PointWorld, and TrackCast (\cref{tab:supp_molmo_tracks}).

The caption describes the context motion in the style of their training captions, as ``a/an \{species\} stands still'' or ``a/an \{species\} \{walks $|$ runs\} \{to the left $|$ to the right $|$ toward the camera $|$ away from the camera\}''.
It is derived from the horizontal motion of the tracks' centroid over the context.
A caption that describes the ground-truth future would use oracle information, so we do not use one.

We fine-tune the released model on the AiM-Track training split (DT4D has no RGB) with the same captions and all points marked visible.
We use their stage-2 recipe with 24 output frames at 10\,fps for 2k steps, by which point the training loss has leveled off.

\expandafter\setlength\csname @fptop\endcsname{0pt}\begin{table}[t]
  \centering
  \caption{\textbf{Error on MolmoMotion's own tracks.} Canonical EPE of \cref{tab:main_results}, restricted to the moving tracks that MolmoMotion forecasts itself, without interpolation (412 of the 429 AiM-Track and 23 of the 24 DAVIS windows have such tracks). EPE in voxels; bold: best, underlined: second.}
  \label{tab:supp_molmo_tracks}
  \small
  \setlength{\tabcolsep}{1.5pt}
  \begin{tabularx}{\columnwidth}{>{\hspace{2pt}}l YY}
    \toprule
    & \multicolumn{2}{c}{Canonical EPE$\downarrow$} \\
    \cmidrule(lr){2-3}
    & AiM-Track & DAVIS \\
    \midrule
    \mname{Persistence} & 5.20 & 6.03 \\
    \mname{PointWorld}~\cite{point_world} & 5.41 & 6.84 \\
    \mname{MolmoMotion (zero-shot)}~\cite{zhang2026molmomotion} & 9.37 & 11.98 \\
    \mname{MolmoMotion (fine-tuned)}~\cite{zhang2026molmomotion} & 6.09 & 8.49 \\
    \rowcolor{ourrow} {\setlength{\tabcolsep}{6pt}\oursbar}\mname{\textbf{TrackCast (1 sample)}} & \second{4.82} & \second{5.36} \\
    \rowcolor{ourrow} {\setlength{\tabcolsep}{6pt}\oursbar}\mname{\textbf{TrackCast (best of 5)}} & \best{4.46} & \best{5.11} \\
    \bottomrule
  \end{tabularx}
\end{table}

\end{document}